\documentclass[11pt,a4paper]{article}
\usepackage[hyperref]{eacl2021}
\usepackage{times}
\usepackage{latexsym}

\usepackage{microtype}

\usepackage{times}
\usepackage{xcolor}
\usepackage{natbib}
\usepackage{soul}
\usepackage[utf8]{inputenc}

\usepackage[nocenter]{qtree}

\usepackage{CJKutf8}
\usepackage{url}
\usepackage{tipa}

\usepackage{amsmath,amsthm,amssymb,amsfonts,bbm,stmaryrd}
\usepackage{enumitem}

\usepackage{textcomp}
\usepackage{multirow} % Allows what
\usepackage{graphicx} % Allows including images
\usepackage{subcaption}
\usepackage{booktabs} % Allows the use of \toprule, \midrule and 
\usepackage[normalem]{ulem}
\usepackage{makecell}
\usepackage{courier}
\usepackage{bbm}
\usepackage{algorithm,algpseudocode}
\usepackage{adjustbox}
\usepackage{tabularx}
\usepackage{setspace}

\usepackage{threeparttable}

\usepackage{todonotes}

\usepackage{listings}
\usepackage{pifont}
\usepackage{color}
\usepackage{bm}
\usepackage{ulem}
\usepackage{array}

\usepackage{pifont}
\newcommand{\mbs}{\boldsymbol}

\newcommand{\mb}{\mathbf}
\newcommand{\bmu}{\bm{\mu}}

\newcommand{\Ra}{\shortrightarrow}

\usepackage{siunitx}
\usepackage{ragged2e}
\newcolumntype{R}[1]{>{\RaggedLeft\arraybackslash}p{#1}}

\newcolumntype{P}[1]{>{\RaggedRight\arraybackslash}p{#1}}

\aclfinalcopy % Uncomment this line for the final submission
\usepackage{arydshln}
\makeatletter
\def\adl@drawiv#1#2#3{%
        \hskip.5\tabcolsep
        \xleaders#3{#2.5\@tempdimb #1{1}#2.5\@tempdimb}%
                #2\z@ plus1fil minus1fil\relax
        \hskip.5\tabcolsep}
\newcommand{\cdashlinelr}[1]{%
  \noalign{\vskip\aboverulesep
           \global\let\@dashdrawstore\adl@draw
           \global\let\adl@draw\adl@drawiv}
  \cdashline{#1}
  \noalign{\global\let\adl@draw\@dashdrawstore
           \vskip\belowrulesep}}
\makeatother

\usepackage{calc}
\usepackage{tikz}
\usepackage{varwidth}
\usepackage{tikz-dependency}
\usetikzlibrary{positioning}
\usetikzlibrary{calc,intersections}
\usetikzlibrary{shapes.multipart}

\definecolor{myorange}{HTML}{FF6700}
\definecolor{mypurple}{HTML}{9900ff}
\definecolor{mygreen}{HTML}{009000}

\makeatletter
\newcommand{\gettikzxy}[3]{%
  \tikz@scan@one@point\pgfutil@firstofone#1\relax
  \edef#2{\the\pgf@x}%
  \edef#3{\the\pgf@y}%
} % https://tex.stackexchange.com/a/321379

\tikzset{state/.style={
		circle,
		minimum size =1.25cm
		draw=black,
		thick,
		fill=orange,
		text=white}}
\tikzset{block/.style={
		rounded corners, fill=blue!15,
		anchor=north west}}
\tikzset{line/.style={
		-latex,
		line width=.5mm,
		black!65}}
\tikzset{annotation/.style={
		align=left,
		anchor=north west,
		draw=black!75,
		minimum height=10mm,
		minimum width=10mm}}
\tikzset{hmodule/.style={
		rounded corners,
		align=center,
		anchor=north west,
		font=\Large,
		fill=blue!20,
		draw=lightgray!75,
		minimum height=10mm,
		minimum width=20mm}}
\tikzset{vmodule/.style={
		rounded corners,
		align=center,
		rotate=90,
		anchor=north west,
		font=\Large,
		fill=blue!20,
		draw=lightgray!75,
		minimum height=10mm,
		minimum width=20mm}}
\tikzset{hvector/.style={
		rounded corners=2.5mm,
		align=center,
		anchor=north west,
		fill=gray!5,
		draw=black!75,
		minimum height=5mm,
		minimum width=20mm}}
\tikzset{vvector/.style={
		rounded corners=2.5mm,
		align=center,
		rotate=90,
		anchor=north west,
		fill=gray!5,
		draw=black!75,
		minimum height=5mm,
		minimum width=20mm}}
\tikzset{square/.style={
    rounded corners=0mm,
    align=center,
    anchor=north west,
    text centered,
    minimum height=10mm,
    minimum width=10mm}}
\tikzset{split_rect/.style={
    rounded corners=0mm,
    align=center,
    anchor=north west,
    rectangle split,
    rectangle split parts=2,
    text centered,
    inner ysep=3.5mm,
    minimum height=20mm,
    minimum width=10mm}}
\usepackage{xspace}
\usepackage[utf8]{inputenc}
\usepackage[T1]{fontenc}    % use 8-bit T1 fonts

\pdfmapline{+delphine < Delphine.ttf <T1-WGL4.enc}
\pdfmapline{+QTWestEndRegular < QTWestEndRegular.ttf <T1-WGL4.enc}

\usepackage{roboto-mono} % \robotomonoRegular
\newcommand*{\qcr}{\fontfamily{qcr}\selectfont}

\usepackage{DejaVuSansMono}
\newcommand\dmonofont[1]{{\smash{\usefont{T1}{DejaVuSansMono-TLF}{m}{n}#1}}}

\newcommand\comicfont[1]{\dmonofont{#1}}
\renewcommand{\tt}{\qcr}  % styles for author names

\newcommand{\dCF}{\comicfont{CF}\xspace}
\newcommand{\dCP}{\comicfont{CP}\xspace}
\newcommand{\dCN}{\comicfont{CN}\xspace}
\newcommand{\dCM}{\comicfont{CM}\xspace}
\newcommand{\dCG}{\comicfont{CG}\xspace}
\newcommand{\dCR}{\comicfont{CR}\xspace}
\newcommand{\dCK}{\comicfont{CK}\xspace}
\newcommand{\dCL}{\comicfont{CL}\xspace}

\newcommand{\dANSWERS}{\comicfont{Answers}\xspace}
\newcommand{\dEMAIL}{\comicfont{Email}\xspace}
\newcommand{\dNEWSGROUP}{\comicfont{Newsgroup}\xspace}
\newcommand{\dREVIEWS}{\comicfont{Reviews}\xspace}
\newcommand{\dWEBLOG}{\comicfont{Weblog}\xspace}

\usepackage{amsmath,amsfonts,bm}

\def\eqref#1{equation~\ref{#1}}
\def\1{\bm{1}}

\def\rvw{{\mathbf{w}}}

\def\rvz{{\mathbf{z}}}

\DeclareMathAlphabet{\mathsfit}{\encodingdefault}{\sfdefault}{m}{sl}
\SetMathAlphabet{\mathsfit}{bold}{\encodingdefault}{\sfdefault}{bx}{n}

\DeclareMathOperator*{\argmax}{arg\,max}

\title{An Empirical Study of Compound PCFGs}

\newcommand{\uoa}{\normalfont \text{\textipa{\ae}}}
\newcommand{\uoe}{\normalfont \text{\textipa{E}}}

\author{Yanpeng Zhao$^{\uoe}$ \\
	$^{\uoe}$ILCC, University of Edinburgh \\
	\tt{yanp.zhao@ed.ac.uk} \\\And
	Ivan Titov$^{\uoe\uoa}$\\
	$^{\uoa}$ILLC, University of Amsterdam \\
	\tt{ititov@inf.ed.ac.uk} \\}

\date{}

\begin{document}
\maketitle
\begin{abstract}
%We present a novel model for transfering phrase-structure syntax from English to Chinese, and vice versa.
%the analysis of the model is mainly central on English treebank.
Compound probabilistic context-free grammars (C-PCFGs) have recently established a new state of the art for unsupervised phrase-structure grammar induction.
However, due to the high space and time complexities of chart-based representation and inference, it is difficult to investigate C-PCFGs comprehensively.
%it is difficult to examine how hyperparameter choices influence the performance (e.g., nonterminal number and embedding dimension).
In this work, we rely on a fast implementation of C-PCFGs to conduct an evaluation complementary to that of~\citet{kim-etal-2019-compound}. We start by analyzing and ablating C-PCFGs on English treebanks.
% : WSJ and Brown of the Penn Treebank, and English Web Treebank.
% , which covers a diverse range of text domains. 
Our findings suggest that (1) C-PCFGs are data-efficient and can generalize to unseen sentence/constituent lengths; and (2) C-PCFGs make the best use of sentence-level information in generating preterminal rule probabilities.
We further conduct a multilingual evaluation of C-PCFGs. The experimental results show that the best configurations of C-PCFGs, which are tuned on English, do not always generalize to morphology-rich languages.
\end{abstract}

\section{Introduction}

Probabilistic context-free grammars (PCFGs) have been used for unsupervised phrase-structure grammar learning since decades ago~\citep{LARI199035},
but learning PCFGs with the Expectation Maximization algorithm~\citep{EM} has been difficult because it involves non-convex optimization.
Recently,~\citet{kim-etal-2019-compound} proposed compound PCFGs,
an over-parameterized neural model that extends corpus-level PCFGs by defining a mixture of PCFGs per sentence.
C-PCFGs not only have achieved state-of-the-art performance on English and Chinese treebanks in the traditional grammar-induction setting but have also been shown to be effective in a visually-grounded learning setting~\citep{zhao-titov-2020-visually}.
However, there still lacks a thorough study of C-PCFGs, largely because of the high space and time complexities of chart-based representation and inference.
% due to the high space- and time-complexity of chart-based representation and inference, it is difficult to study C-PCFGs comprehensively.

In this work, we rely on a fast implementation
% {\url{https://git.io/JJAfu}}
of C-PCFGs to conduct a set of experiments complementary to those of~\citet{kim-etal-2019-compound}.\footnote {\href{https://github.com/zhaoyanpeng/cpcfg}{https://github.com/zhaoyanpeng/cpcfg}.}
Our first study focuses on data efficiency and length generalization of C-PCFGs. We conduct experiments on three English treebanks: the Wall Street Journal (WSJ) and Brown portions of the Penn Treebank~\citep{ptree} and the English Web Treebank~\citep{enweb}, which cover fourteen text domains, including news, biography, fiction, and web text. 
We empirically find that, though trained only on short sentences, C-PCFGs can generalize to longer sentences while maintaining high performance at test time. 
For example, a C-PCFG that is trained only on WSJ training sentences shorter than 31 tokens achieves 54.8\% F1 on the full WSJ test set. This, on the other hand, demonstrates that C-PCFGs are data-efficient.
%This suggests that C-PCFGs are data efficiency.
%They can uncover hierarchical structures of sentences from a small set of easy data.

We further investigate what factors contribute to the improved performance of C-PCFGs over traditional PCFGs. 
Since the major difference between C-PCFGs and classical PCFGs is that C-PCFGs define sentence-dependent rule probabilities by using global sentence-level information, %an additional sentence embedding,
we ablate C-PCFGs by individually removing this information from their three types of rule: start, nonterminal, and preterminal rules.\footnote{Start rules generate a nonterminal symbol from the start symbol $S$ (e.g., $S\Ra A$), preterminal rules generate a word from a nonterminal symbol (e.g., $A\Ra w$), and nonterminal rules are binary rules of the form $A\Ra BC$, which involve only nonterminal symbols.}
We conduct ablation studies on the three English treebanks. Our experimental results show that sentence-level information is most effective for preterminal rules.
%presumably because it encodes part-of-speech tags which are important for parsing.

Despite being performant on English,
it is still unclear whether C-PCFGs can generalize to languages beyond English, so we further conduct a multilingual evaluation of C-PCFGs on nine additional languages, including Chinese from the Chinese Penn Treebank~\citep{ctb} and the other eight languages (Basque, German, French, etc.) from SPMRL~\citep{seddah-etal-2014-introducing}.
Our findings suggest that the best configurations of C-PCFGs, which are tuned on English WSJ,  do not necessarily generalize to morphology-rich languages.
%and find that they can generalize to languages beyond English.

%We organize the rest of paper as follows:
%we briefly describe the background of C-PCFGs in Section~\ref{sec:background}.
%Section~\ref{sec:setup} summarizes datasets, evaluation methods, and implementation details.
%In Section~\ref{sec:ablation} we present comprehensive evaluation and analysis of C-PCFGs.

\section{Compound PCFGs}\label{sec:background}

Compound PCFGs adopt a novel parameterization of PCFGs.
Unlike PCFGs, which assign each grammar rule $r$ a non-negative \text{scalar} $\pi_r$ such that
$\sum_{r: A\Ra \gamma} \pi_r = 1$ for each given left-hand-side symbol $A$ ($\gamma$ indicates any grammar symbols),
C-PCFGs relax the strong context-free assumption of PCFGs by
assuming that rule probabilities follow a compound distribution:
\begin{flalign}
\pi_r = g_r(\rvz; \theta) \quad \text{with} \quad \rvz\sim p(\rvz)\,,
\end{flalign}
where $p(\rvz)$ is a prior distribution and allows for capturing interdependencies among grammar rules;
$g_{r}(\rvz; \theta)$ generates the rule probability conditioning on the the latent $\rvz$.
Typically, $g_{r}(\rvz; \theta)$ is parameterized by flexible neural networks and is amenable to gradient-based optimization (we refer interested readers to~\citet{kim-etal-2019-compound} for the details on parameterization).

The learning of C-PCFGs is formulated as maximizing the log-likelihood of each observed sentence $\mbs{w} = w_1w_2\ldots w_n$:
\begin{flalign}
%\mathcal{L}(\boldsymbol{w}; \theta) = -
\log p_{\theta}(\boldsymbol{w}) = \log
\int_{\rvz} \sum_{t\in \mathcal{T}_{\mathcal{G}}(\boldsymbol{w})} p_{\theta}(t | \rvz) p(\rvz)\,d\rvz\,, 
\end{flalign}
where $\mathcal{T}_{\mathcal{G}}(\boldsymbol{w})$ consists of all possible parses of the sentence $\mbs{w}$ under a PCFG $\mathcal{G}$.
As standard in learning latent variable models,
C-PCFGs resort to variational inference for tractable learning
and, instead, maximize the evidence lower bound (ELBO):
\begin{flalign}\label{eq:cpcfg_elbo}
&\log p_{\theta}(\boldsymbol{w}) \geq \text{ELBO}(\boldsymbol{w}; \phi, \theta) = \\
&\mathbb{E}_{q_{\phi}(\rvz | \boldsymbol{w})}[\log p_{\theta}(\boldsymbol{w} | \rvz)] - 
\text{KL}[q_{\phi}(\rvz | \boldsymbol{w}) || p(\rvz)]\,, \nonumber
\end{flalign}
where the first term computes the expected log-likelihood under a variational posterior $q_{\phi}(\rvz | \mbs{w})$, which defines a distribution over the latent $\rvz$ and is parameterized by a neural network.
The second term, i.e., the Kullback–Leibler (KL) divergence, can be estimated analytically when $p(\rvz)$ and $q_{\phi}(\rvz | \mbs{w})$ are normally distributed.

For each given $\rvz$, C-PCFGs satisfy the context-free assumption and thus admit tractable inference.
At inference time, we seek the most probable parse $t^*$ of $\mbs{w}$:
\begin{flalign}
t^{*} = \argmax\int_{\rvz} p_{\theta}(t | \rvw, \rvz) p_{\theta}(\rvz | \boldsymbol{w})\,d\rvz\,.
\end{flalign}
Though given $\rvz$, the maximum a posterior (MAP) inference over $p_{\theta}(t | \rvw, \rvz)$ can be exactly solved by using the CKY algorithm~\citep{cocke1969programming,kasami1966efficient,YOUNGER1967189},
the integral over $\rvz$ renders inference intractable.
%However, the intergral over $\rvz$ renders inference intractable.
Instead, the MAP inference is approximated by:
\begin{flalign}
\!\!\!\!
t^{*} \approx \argmax\int_{\rvz} p_{\theta}(t | \rvw, \rvz) \delta(\rvz - \bmu_{\phi}(\boldsymbol{w}))\,d\rvz\,,
\end{flalign}
where $\delta(\cdot)$ is the Dirac delta function
and $\bmu_{\phi}(\boldsymbol{w})$ is the mean vector of the variational posterior.

Similarly to C-PCFGs, neural PCFGs (N-PCFGs) also use neural networks to parameterize PCFGs,
%The neural networks they used have a similar form to those in C-PCFGs,
but their parameterization does not rely on the sentence-dependent $\mb{z}$.
%Thus, from the modeling perspective, 
%a major difference between C-PCFGs and N-PCFGs is the use of sentence-level text features $\mb{z}$.
In the following discussion, we will refer to $\rvz$ as ``sentence embedding."

\begin{table*}[t]\small
\centering
{\setlength{\tabcolsep}{.65em}
\makebox[\linewidth]{\resizebox{\linewidth}{!}{%
\begin{tabular}{rllllllll}
    \toprule
    \multicolumn{1}{r}{\textbf{Model}} &
    \multicolumn{1}{l}{\textbf{NP}} & 
    \multicolumn{1}{l}{\textbf{VP}} & 
    \multicolumn{1}{l}{\textbf{PP}} & 
    \multicolumn{1}{l}{\textbf{SBAR}} & 
    \multicolumn{1}{l}{\textbf{ADJP}} & 
    \multicolumn{1}{l}{\textbf{ADVP}} & 
    \multicolumn{1}{l}{\textbf{C-F1}} & 
    \multicolumn{1}{l}{\textbf{S-F1}} \\
    \midrule
    Left Branching & 10.4 & \phantom{0}0.5 & \phantom{0}5.0 & \phantom{0}5.3 & \phantom{0}2.5 & \phantom{0}8.0 & \phantom{0}6.0 & \phantom{0}8.7 \\
    Right Branching & 24.1 & \textbf{71.5} & 42.4 & \textbf{68.7} & 27.7 & 38.1 & 36.1 & 39.5 \\
    Random Trees & 22.5$_{\pm 0.3}$ & 12.3$_{\pm 0.3}$ & 19.0$_{\pm 0.5}$ & \phantom{0}9.3$_{\pm 0.6}$  & 24.3$_{\pm 1.7}$ & 26.9$_{\pm 1.3}$ & 15.3$_{\pm 0.1}$ & 18.1$_{\pm 0.1}$ \\
    \midrule
    $^\dagger$N-PCFG  & 71.2 & 33.8 & 58.8 & 52.5 & 32.5 & 45.5  &   & 50.8   \\
    N-PCFG & 72.2$_{\pm 4.8}$ & 31.4$_{\pm 9.7}$ & 66.8$_{\pm 4.7}$ & 50.2$_{\pm 9.1}$ & 46.3$_{\pm 5.7}$ & 55.2$_{\pm 5.0}$ & 49.0$_{\pm 3.5}$ & 50.8$_{\pm 3.8}$ \\
    $^\dagger$C-PCFG  & 74.7 & 41.7 & 68.8 & 56.1 & 40.4 & 52.5  &   & 55.2   \\
    %					C-PCFG & 75.9$_{\pm 3.6}$ & 40.8$_{\pm 3.3}$ & \textbf{72.3}$_{\pm 0.6}$ & 60.8$_{\pm 5.8}$ & \textbf{47.5}$_{\pm 6.0}$ & 63.5$_{\pm 3.5}$ & \textbf{53.8}$_{\pm 1.9}$ & \textbf{56.0}$_{\pm 1.8}$ \\
    \cdashlinelr{2-9}
    C-PCFG & 76.7$_{\pm 2.0}$ & 40.7$_{\pm 5.5}$ & 71.3$_{\pm 2.1}$ & 53.8$_{\pm 3.1}$ & 45.9$_{\pm 2.8}$ & 64.2$_{\pm 2.8}$ & 53.5$_{\pm 1.4}$ & 55.7$_{\pm 1.3}$ \\
    w/ shared \dmonofont{R} & 74.2$_{\pm 3.2}$ & 38.3$_{\pm 6.6}$ & 68.0$_{\pm 5.1}$ & 55.3$_{\pm 4.5}$ & 43.3$_{\pm 13.1}$ & 57.6$_{\pm 7.9}$ & 51.7$_{\pm 3.5}$ & 53.7$_{\pm 3.5}$ \\
    w/ shared \dmonofont{N} & 73.0$_{\pm 1.4}$ & 35.2$_{\pm 5.6}$ & 70.4$_{\pm 2.3}$ & 51.7$_{\pm 5.6}$ & 41.3$_{\pm 11.5}$ & 54.2$_{\pm 7.5}$ & 50.6$_{\pm 1.8}$ & 53.0$_{\pm 1.8}$ \\
    w/ shared \dmonofont{P} & 74.0$_{\pm 5.7}$ & 26.1$_{\pm 4.1}$ & 69.7$_{\pm 1.0}$ & 51.7$_{\pm 6.1}$ & 37.8$_{\pm 8.3}$ & 54.1$_{\pm 6.8}$ & 49.0$_{\pm 2.4}$ & 50.7$_{\pm 1.8}$ \\
    \midrule
    L50C-PCFG & \textbf{76.9}$_{\pm 3.6}$ & 40.7$_{\pm 3.7}$ & \textbf{72.3}$_{\pm 0.6}$ & 60.1$_{\pm 5.5}$ & \textbf{46.9}$_{\pm 5.8}$ & 63.2$_{\pm 5.0}$ & \textbf{53.8}$_{\pm 2.1}$ & \textbf{55.9}$_{\pm 1.9}$ \\
    %					L40: C-PCFG & 75.9$_{\pm 3.6}$ & 40.8$_{\pm 3.3}$ & \textbf{72.3}$_{\pm 0.6}$ & 60.8$_{\pm 5.8}$ & \textbf{47.5}$_{\pm 6.0}$ & 63.5$_{\pm 3.5}$ & \textbf{53.8}$_{\pm 1.9}$ & \textbf{56.0}$_{\pm 1.8}$ \\
    L40C-PCFG & 76.7$_{\pm 2.0}$ & 40.7$_{\pm 5.5}$ & 71.3$_{\pm 2.1}$ & 53.8$_{\pm 3.1}$ & 45.9$_{\pm 2.8}$ & 64.2$_{\pm 2.8}$ & 53.5$_{\pm 1.4}$ & 55.7$_{\pm 1.3}$ \\
    L30C-PCFG & 74.5$_{\pm 2.8}$ & 38.4$_{\pm 1.7}$ & 71.1$_{\pm 1.2}$ & 59.7$_{\pm 4.8}$ & 44.2$_{\pm 4.1}$ & \textbf{64.3}$_{\pm 3.1}$ & 52.5$_{\pm 1.5}$ & 54.8$_{\pm 1.4}$ \\
    L20C-PCFG & 72.4$_{\pm 2.3}$ & 36.5$_{\pm 1.1}$ & 69.2$_{\pm 1.7}$ & 54.1$_{\pm 3.2}$ & 41.9$_{\pm 2.3}$ & 58.1$_{\pm 7.1}$ & 50.6$_{\pm 0.9}$ & 52.8$_{\pm 0.7}$ \\
    L10C-PCFG & 67.1$_{\pm 3.8}$ & 31.0$_{\pm 9.8}$ & 61.3$_{\pm 2.2}$ & 45.9$_{\pm 8.2}$ & 36.7$_{\pm 2.3}$ & 41.3$_{\pm 6.0}$ & 45.5$_{\pm 2.4}$ & 48.2$_{\pm 2.3}$ \\
    % \midrule
    % C-PCFG & 76.7$_{\pm 2.0}$ & 40.7$_{\pm 5.5}$ & 71.3$_{\pm 2.1}$ & 53.8$_{\pm 3.1}$ & 45.9$_{\pm 2.8}$ & 64.2$_{\pm 2.8}$ & 53.5$_{\pm 1.4}$ & 55.7$_{\pm 1.3}$ \\
    % w/ shared \dmonofont{R} & 74.2$_{\pm 3.2}$ & 38.3$_{\pm 6.6}$ & 68.0$_{\pm 5.1}$ & 55.3$_{\pm 4.5}$ & 43.3$_{\pm 13.1}$ & 57.6$_{\pm 7.9}$ & 51.7$_{\pm 3.5}$ & 53.7$_{\pm 3.5}$ \\
    % w/ shared \dmonofont{N} & 73.0$_{\pm 1.4}$ & 35.2$_{\pm 5.6}$ & 70.4$_{\pm 2.3}$ & 51.7$_{\pm 5.6}$ & 41.3$_{\pm 11.5}$ & 54.2$_{\pm 7.5}$ & 50.6$_{\pm 1.8}$ & 53.0$_{\pm 1.8}$ \\
    % w/ shared \dmonofont{P} & 74.0$_{\pm 5.7}$ & 26.1$_{\pm 4.1}$ & 69.7$_{\pm 1.0}$ & 51.7$_{\pm 6.1}$ & 37.8$_{\pm 8.3}$ & 54.1$_{\pm 6.8}$ & 49.0$_{\pm 2.4}$ & 50.7$_{\pm 1.8}$ \\
    \bottomrule
\end{tabular}}}}
\caption{\label{tab:test_wsj}
Recall on six frequent constituent labels (NP, VP, PP, SBAR, ADJP, ADVP) in the WSJ test data, corpus-level F1 (C-F1), and sentence-level F1 (S-F1) results.
The best mean number in each column is in bold.
$\dagger$ denotes results reported by~\citet{kim-etal-2019-compound}.
L\# indicates that the models are trained on sentences no longer than \# tokens.
}
\end{table*}
\begin{figure*}[t]
\centering
\begin{subfigure}{.33\linewidth}
    \includegraphics[width=1.\linewidth]{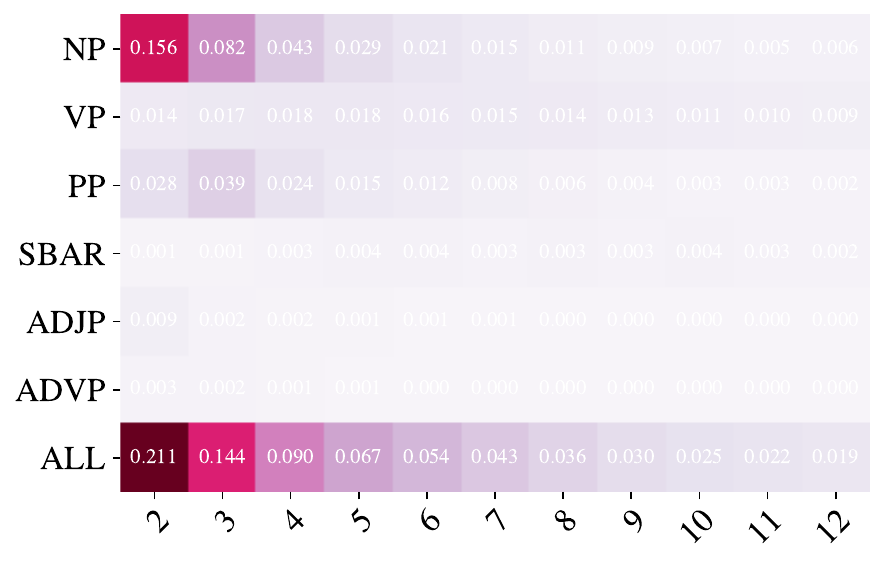}
    \caption{WSJ Test Set.}
    \label{fig:lspan-label-wsj-test}
\end{subfigure}
\hfill
\begin{subfigure}{.33\linewidth}
    \includegraphics[width=1.\linewidth]{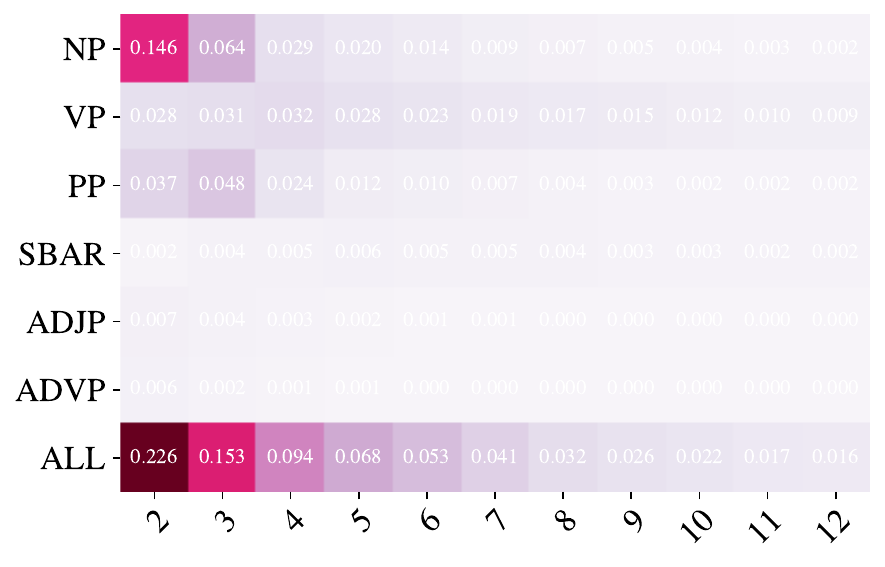}
    \caption{Brown Test Set.}
    \label{fig:lspan-label-brown-test}
\end{subfigure}%
\hfill
\begin{subfigure}{.33\linewidth}
    \includegraphics[width=1.\linewidth]{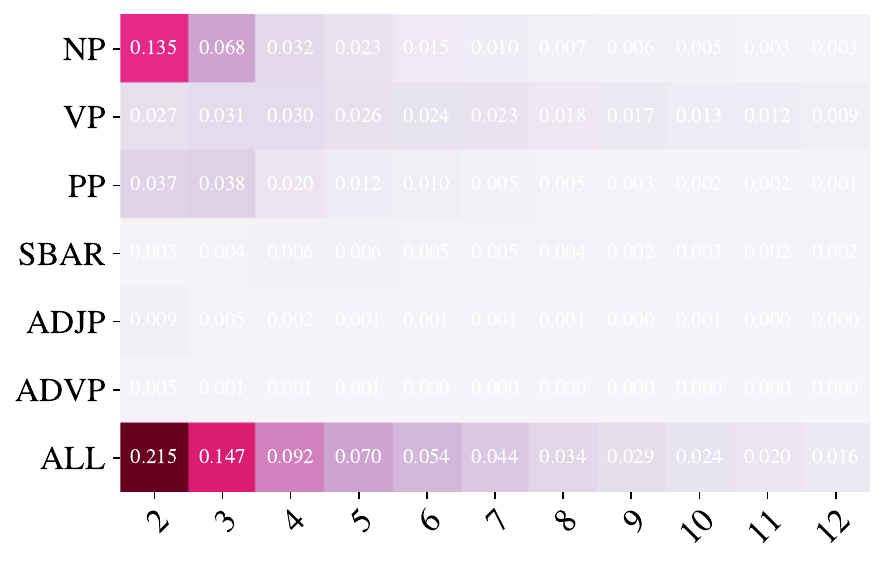}
    \caption{Enweb Test Set.}
    \label{fig:lspan-label-enweb-test}
\end{subfigure}%
% \\
% \vspace{2mm}
% \begin{subfigure}{.33\linewidth}
%     \includegraphics[width=1.\linewidth]{figure/recall_by_len/recall_by_len_test_wsj.pdf}
%     \caption{WSJ Test Set.}
%     \label{fig:lspan-label-wsj-test}
% \end{subfigure}
% \hfill
% \begin{subfigure}{.33\linewidth}
%     \includegraphics[width=1.\linewidth]{figure/recall_by_len/recall_by_len_test_brown.pdf}
%     \caption{Brown Test Set.}
%     \label{fig:lspan-label-brown-test}
% \end{subfigure}%
% \hfill
% \begin{subfigure}{.33\linewidth}
%     \includegraphics[width=1.\linewidth]{figure/recall_by_len/recall_by_len_test_enweb.pdf}
%     \caption{Enweb Test Set.}
%     \label{fig:lspan-label-enweb-test}
% \end{subfigure}%
\caption{\label{fig:label_dist_per_length}
Phrasal label distribution broken down per constituent length.
}
\end{figure*}

\section{Experimental setup}\label{sec:setup}

%\subsection{Datasets and evaluation}
% \noindent\textbf{Datasets:}
We investigate C-PCFGs across ten languages but primarily focus on English. 

\paragraph{English Treebanks.} For English, we consider three treebanks: the Wall Street Journal (WSJ) and Brown portions of the Penn Treebank~\citep{marcus-etal-1994-penn} and the English Web Treebank (Enweb;~\citet{enweb}). 
Since Brown and Enweb are composed of eight and five subdomains, respectively, we will be actually conducting experiments on fourteen domains.
Below we elaborate on Brown and Enweb since they are not as widely used as WSJ.

\begin{itemize}[wide=0\parindent]
\item \emph{Brown} is part of the Penn Treebank and consists of manually parsed sentences from eight domains, including lore, biography, fiction, and humor~\citep{ptree}. We divide sentences in each domain into three splits: approximately 70\% of the sentences for training, 15\% for development, and 15\% for test. We further merge the training, development, and test subsets across domains and create a mixed-domain Brown.
\item \emph{Enweb} stands for the English Web Treebank and consists of sentences from five domains: weblogs, newsgroups, email, reviews, and question-answers~\citep{enweb}. 
% It is created with the goal of evaluating the robustness of parsing models across domains and thus is ideal for testing our transfer learning models. 
Each domain contains sentences that have been manually annotated with phrase structures. We divide sentences in each domain in a similar way to that we divide Brown sentences. We also create a mixed-domain Enweb. 
\end{itemize}

\paragraph{Multilingual Treebanks.} We conducted a multilingual evaluation on the rest nine languages, including Chinese, Basque, German, French, Hebrew, Hungarian, Korean, Polish, and Swedish.
For Chinese, we use the Penn Chinese Treebank 5.1 (CTB;~\citet{ctb}).
For the other eight languages, we rely on their treebanks from the SPMRL 2014 shared task~\citep{seddah-etal-2014-introducing}.

We use the standard data splits for each treebank whenever there have been established standards for data set splitting. %(see Section~\ref{sec:exp_mul}).
Following~\citet{kim-etal-2019-compound}, we remove punctuation from all the treebanks, and keep the top 10,000 frequent tokens in the training split of each treebank as the vocabulary.\footnote{A unified data preprocessing pipeline is available at \href{https://github.com/zhaoyanpeng/xcfg}{https://github.com/zhaoyanpeng/xcfg}.}
Unless otherwise specified, we train C-PCFGs on sentences no longer than 40 tokens, namely L40C-PCFG.

\paragraph{Model hyperparameters and evaluation:} 
We re-implement C-PCFGs relying on Torch-Struct~\citep{rush-2020-torch} and adopt the same hyperparameter settings as in~\citet{kim-etal-2019-compound}.
We train C-PCFGs for each language separately.
On each treebank we run C-PCFGs four times with different random seeds and for 30 epochs.
The best model in each run is selected according to the perplexity of the development data.
At test time, trivial spans, such as single-word and sentence-level spans, are ignored.
We report average corpus- and sentence-level F1 numbers as well as unbiased standard deviations.

\begin{figure*}[t]
\centering
% \begin{subfigure}{.33\linewidth}
%     \includegraphics[width=1.\linewidth]{figure/lspan_label_dist_wsj_test.pdf}
%     % \caption{WSJ Test Set.}
%     \label{fig:lspan-label-wsj-test}
% \end{subfigure}
% \hfill
% \begin{subfigure}{.33\linewidth}
%     \includegraphics[width=1.\linewidth]{figure/lspan_label_dist_brown_test.pdf}
%     % \caption{Brown Test Set.}
%     \label{fig:lspan-label-brown-test}
% \end{subfigure}%
% \hfill
% \begin{subfigure}{.33\linewidth}
%     \includegraphics[width=1.\linewidth]{figure/lspan_label_dist_enweb_test.pdf}
%     % \caption{Enweb Test Set.}
%     \label{fig:lspan-label-enweb-test}
% \end{subfigure}%
% \\
% \vspace{2mm}
\begin{subfigure}{.33\linewidth}
    \includegraphics[width=1.\linewidth]{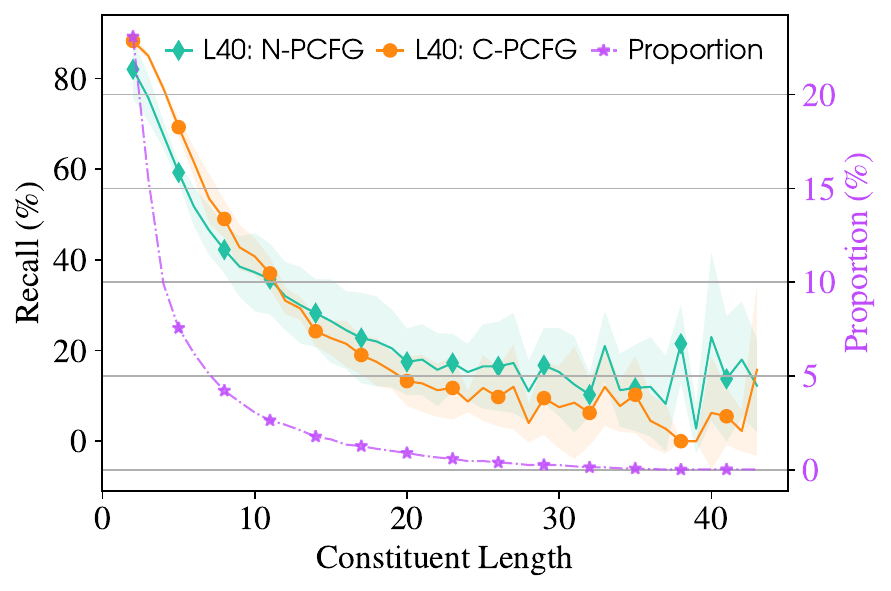}
    \caption{WSJ Test Set.}
    \label{fig:lspan-recall-wsj-test}
\end{subfigure}
\hfill
\begin{subfigure}{.33\linewidth}
    \includegraphics[width=1.\linewidth]{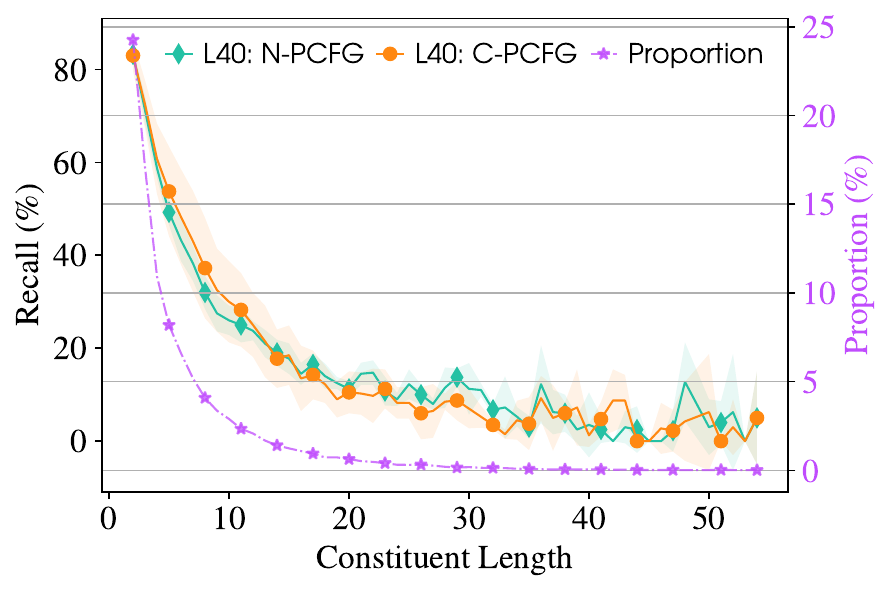}
    \caption{Brown Test Set.}
    \label{fig:lspan-recall-brown-test}
\end{subfigure}%
\hfill
\begin{subfigure}{.33\linewidth}
    \includegraphics[width=1.\linewidth]{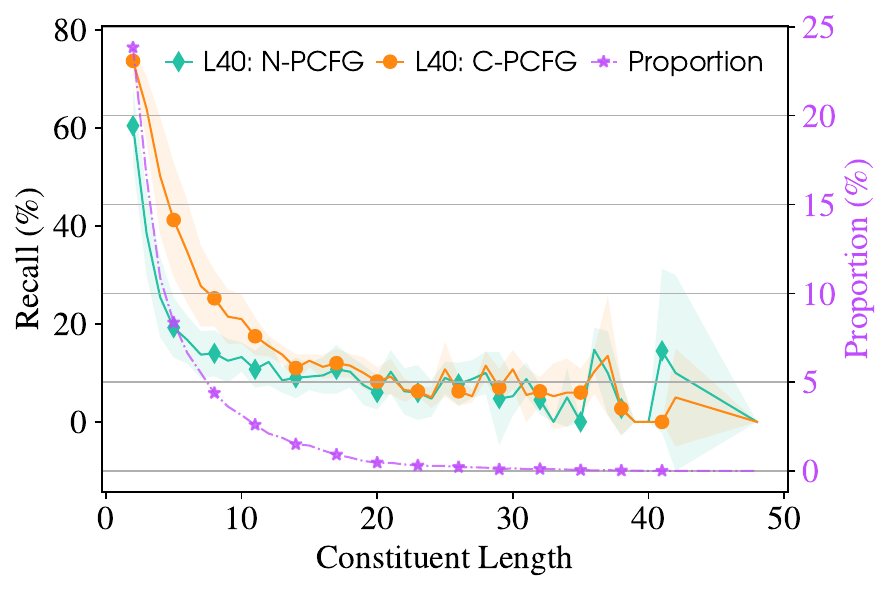}
    \caption{Enweb Test Set.}
    \label{fig:lspan-recall-enweb-test}
\end{subfigure}%
\caption{\label{fig:recall_per_length_test}
Recall broken down per constituent length.
}
\end{figure*}
\begin{figure*}[t!]
\centering
\begin{subfigure}{.33\linewidth}
    \includegraphics[width=1.\linewidth]{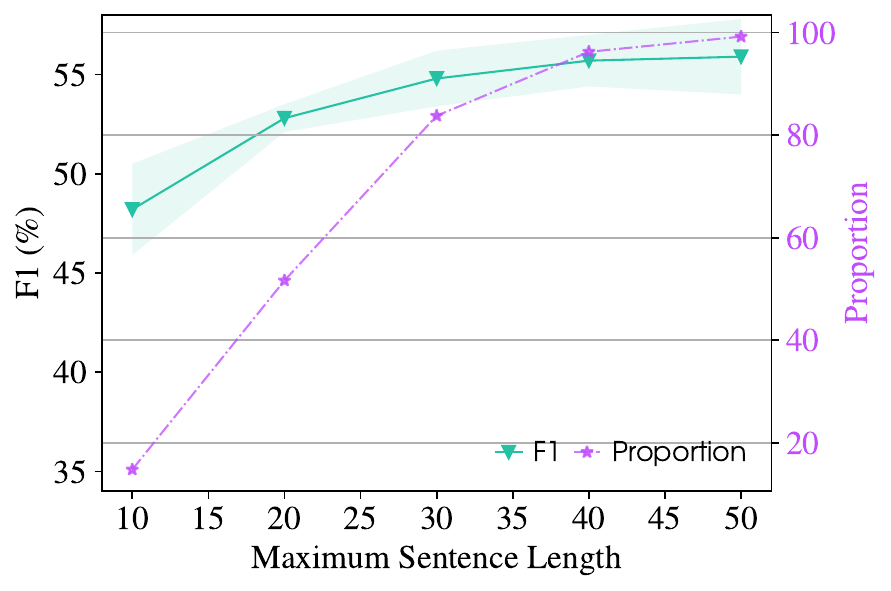}
    \caption{WSJ Test Set.}
    \label{fig:f1-mlen-wsj-test}
\end{subfigure}
\hfill
\begin{subfigure}{.33\linewidth}
    \includegraphics[width=1.\linewidth]{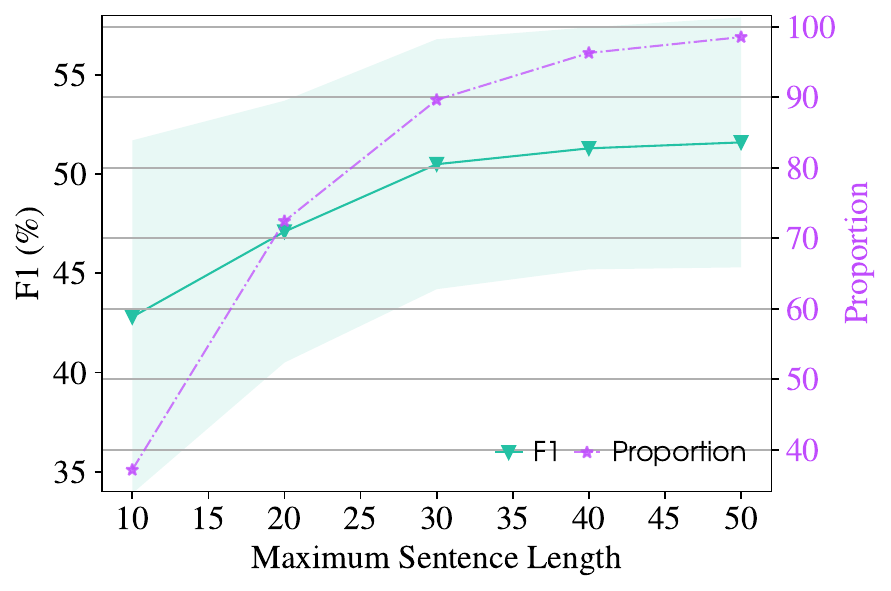}
    \caption{Brown Test Set.}
    \label{fig:f1-mlen-brown-test}
\end{subfigure}%
\hfill
\begin{subfigure}{.33\linewidth}
    \includegraphics[width=1.\linewidth]{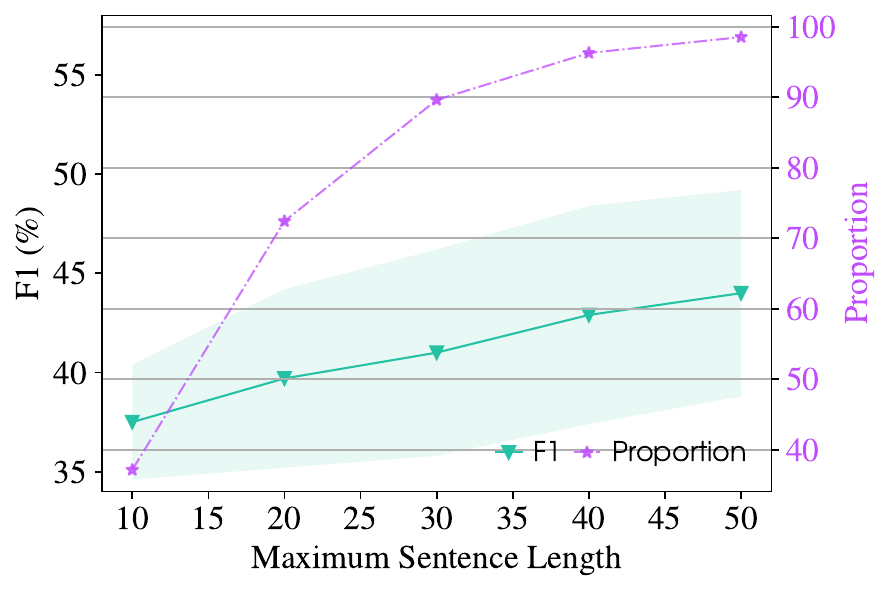}
    \caption{Enweb Test Set.}
    \label{fig:f1-mlen-enweb-test}
\end{subfigure}%
\caption{\label{fig:efficiency-final}
F1 numbers on the three test sets with varying maximum lengths of training sentences.
}
\end{figure*}
\begin{figure*}[t!]
\centering
\includegraphics[width=1.\linewidth]{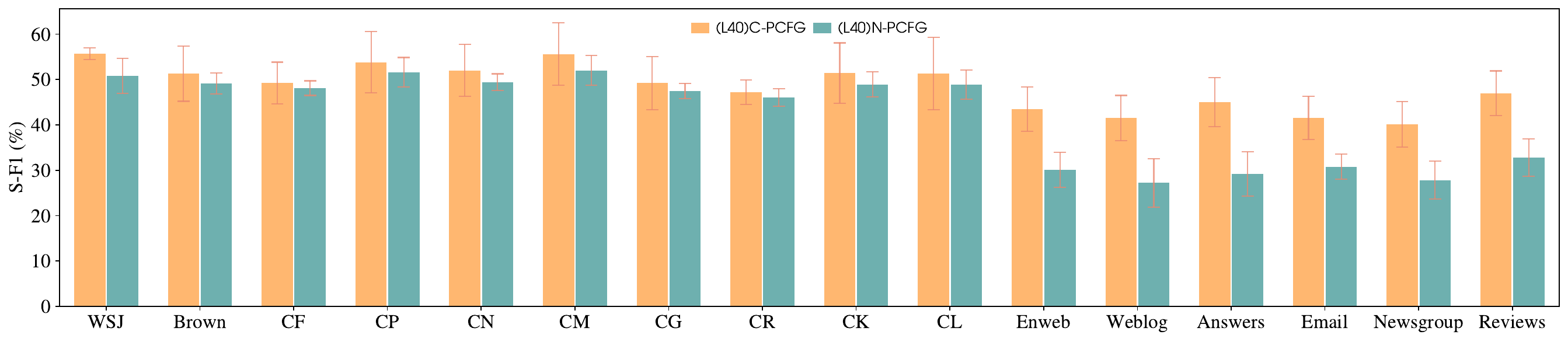}
\caption{\label{fig:ood-subdomain}
Sentence-level F1 on WSJ, mixed-domain Brown and Enweb, and subdomains of Brown and Enweb.
}
\vspace{-.5em}
\end{figure*}

\begin{table*}[t]\small
\centering
{\setlength{\tabcolsep}{.65em}
\makebox[\linewidth]{\resizebox{\linewidth}{!}{%
\begin{tabular}{rllllllll}
    \toprule
    \multicolumn{1}{r}{\textbf{Model}} &
    \multicolumn{1}{l}{\textbf{NP}} & 
    \multicolumn{1}{l}{\textbf{VP}} & 
    \multicolumn{1}{l}{\textbf{PP}} & 
    \multicolumn{1}{l}{\textbf{SBAR}} & 
    \multicolumn{1}{l}{\textbf{ADJP}} & 
    \multicolumn{1}{l}{\textbf{ADVP}} & 
    \multicolumn{1}{l}{\textbf{C-F1}} & 
    \multicolumn{1}{l}{\textbf{S-F1}} \\
    \midrule
    Left Branching & \phantom{0}7.9 & \phantom{0}0.7 & \phantom{0}3.9 & \phantom{0}7.0 & \phantom{0}3.1 & 15.2 & \phantom{0}5.2 & \phantom{0}8.3 \\
    Right Branching & 24.9 & \textbf{65.0} & 38.7 & \textbf{58.6} & 31.6 & 20.4 & 37.1 & 45.3 \\
    Random Trees & 24.7$_{\pm 0.2}$ & 15.0$_{\pm 0.2}$ & 21.3$_{\pm 0.6}$ & 11.7$_{\pm 1.3}$ & 22.1$_{\pm 0.9}$ & 28.9$_{\pm 3.3}$ & 16.5$_{\pm 0.2}$ & 21.2$_{\pm 0.2}$ \\
    \midrule
    N-PCFG & 73.9$_{\pm 1.3}$ & 26.6$_{\pm 5.1}$ & 70.6$_{\pm 1.2}$ & 51.3$_{\pm 4.5}$ & 43.7$_{\pm 4.5}$ & 57.8$_{\pm 1.1}$ & 46.3$_{\pm 1.8}$ & 49.1$_{\pm 2.3}$ \\
    \cdashlinelr{2-9}
    \multicolumn{9}{c}{Per-domain Performance of N-PCFG} \\
    \cdashlinelr{2-9}
\dCF & 70.1$_{\pm 1.7}$ & 21.9$_{\pm 3.9}$ & 67.1$_{\pm 2.5}$ & 47.4$_{\pm 4.9}$ & 44.2$_{\pm 5.0}$ & 53.5$_{\pm 4.7}$ & 45.5$_{\pm 1.7}$ & 48.1$_{\pm 1.6}$ \\
\dCP & 77.5$_{\pm 1.3}$ & 29.7$_{\pm 7.6}$ & 75.9$_{\pm 0.6}$ & 56.4$_{\pm 7.2}$ & 46.8$_{\pm 4.9}$ & 58.2$_{\pm 2.4}$ & 48.2$_{\pm 2.7}$ & 51.6$_{\pm 3.3}$ \\
\dCN & 77.9$_{\pm 1.4}$ & 29.8$_{\pm 4.3}$ & 74.1$_{\pm 1.5}$ & 54.4$_{\pm 4.1}$ & 47.8$_{\pm 3.4}$ & 52.4$_{\pm 7.8}$ & 47.8$_{\pm 1.8}$ & 49.4$_{\pm 1.8}$ \\
\dCM & 80.3$_{\pm 0.3}$ & 29.8$_{\pm 7.1}$ & 82.4$_{\pm 2.6}$ & 53.7$_{\pm 7.1}$ & 52.1$_{\pm 4.2}$ & 56.2$_{\pm 12.5}$ & 51.0$_{\pm 2.2}$ & 52.0$_{\pm 3.3}$ \\
\dCG & 72.0$_{\pm 1.2}$ & 24.0$_{\pm 4.8}$ & 68.3$_{\pm 1.2}$ & 48.6$_{\pm 4.6}$ & 37.1$_{\pm 6.8}$ & 60.5$_{\pm 10.5}$ & 44.9$_{\pm 1.0}$ & 47.5$_{\pm 1.7}$ \\
\dCR & 69.0$_{\pm 1.3}$ & 23.5$_{\pm 4.5}$ & 65.3$_{\pm 1.8}$ & 46.7$_{\pm 6.4}$ & 41.2$_{\pm 5.7}$ & 59.5$_{\pm 8.1}$ & 44.4$_{\pm 1.8}$ & 46.0$_{\pm 1.9}$ \\
\dCK & 73.9$_{\pm 2.0}$ & 27.2$_{\pm 4.4}$ & 69.1$_{\pm 0.8}$ & 49.1$_{\pm 3.8}$ & 42.2$_{\pm 4.3}$ & 63.3$_{\pm 2.1}$ & 45.2$_{\pm 2.1}$ & 48.9$_{\pm 2.8}$ \\
\dCL & 77.9$_{\pm 1.5}$ & 28.8$_{\pm 6.8}$ & 74.5$_{\pm 1.3}$ & 58.6$_{\pm 1.3}$ & 48.6$_{\pm 7.5}$ & 56.0$_{\pm 4.0}$ & 46.8$_{\pm 2.7}$ & 48.9$_{\pm 3.2}$ \\
    \cdashlinelr{1-9}   
    C-PCFG & 75.0$_{\pm 3.1}$ & 31.9$_{\pm 16.2}$ & 67.2$_{\pm 8.5}$ & 54.6$_{\pm 3.9}$ & 39.7$_{\pm 7.8}$ & 59.4$_{\pm 2.6}$ & 47.8$_{\pm 4.4}$ & 51.3$_{\pm 6.1}$ \\
    w/ shared \dmonofont{R} & 73.9$_{\pm 1.2}$ & 30.5$_{\pm 11.8}$ & 71.8$_{\pm 1.2}$ & 52.6$_{\pm 2.7}$ & 38.7$_{\pm 4.5}$ & 58.1$_{\pm 1.7}$ & 47.6$_{\pm 2.8}$ & 50.7$_{\pm 4.3}$ \\
    w/ shared \dmonofont{N} & 74.5$_{\pm 2.0}$ & 38.1$_{\pm 20.4}$ & 70.9$_{\pm 0.8}$ & 56.3$_{\pm 4.7}$ & \textbf{45.4}$_{\pm 9.4}$ & 57.7$_{\pm 2.9}$ & \textbf{49.5}$_{\pm 5.7}$ & 53.1$_{\pm 7.0}$ \\
    w/ shared \dmonofont{P} & 74.4$_{\pm 0.6}$ & 24.2$_{\pm 4.6}$ & \textbf{74.0}$_{\pm 1.4}$ & 56.8$_{\pm 2.2}$ & 41.3$_{\pm 5.3}$ & \textbf{60.0}$_{\pm 2.5}$ & 46.4$_{\pm 0.5}$ & 48.6$_{\pm 1.4}$ \\
    \cdashlinelr{2-9}
    \multicolumn{9}{c}{Per-domain Performance of C-PCFG} \\
    \cdashlinelr{2-9}   
\dCF & 71.2$_{\pm 4.0}$ & 26.8$_{\pm 14.2}$ & 63.2$_{\pm 8.6}$ & 50.9$_{\pm 6.8}$ & 42.9$_{\pm 9.6}$ & 56.7$_{\pm 7.7}$ & 46.8$_{\pm 3.7}$ & 49.2$_{\pm 4.6}$ \\
\dCP & 78.6$_{\pm 3.0}$ & 34.3$_{\pm 17.9}$ & 71.7$_{\pm 8.1}$ & 60.1$_{\pm 4.6}$ & 39.5$_{\pm 5.5}$ & 59.6$_{\pm 7.9}$ & 49.4$_{\pm 5.1}$ & 53.8$_{\pm 6.8}$ \\
\dCN & 80.0$_{\pm 3.0}$ & 35.9$_{\pm 15.5}$ & 72.0$_{\pm 9.1}$ & 54.6$_{\pm 4.6}$ & 44.8$_{\pm 8.4}$ & 60.4$_{\pm 4.9}$ & 50.1$_{\pm 4.4}$ & 52.0$_{\pm 5.7}$ \\
\dCM & 80.6$_{\pm 1.3}$ & 37.6$_{\pm 19.3}$ & 78.6$_{\pm 13.2}$ & 61.8$_{\pm 6.7}$ & 50.0$_{\pm 13.2}$ & 46.9$_{\pm 12.0}$ & 53.6$_{\pm 4.4}$ & 55.6$_{\pm 6.9}$ \\
\dCG & 72.7$_{\pm 4.0}$ & 29.3$_{\pm 16.2}$ & 63.8$_{\pm 8.4}$ & 52.0$_{\pm 3.0}$ & 35.7$_{\pm 7.2}$ & 58.3$_{\pm 2.2}$ & 46.2$_{\pm 4.5}$ & 49.2$_{\pm 5.9}$ \\
\dCR & 69.9$_{\pm 2.9}$ & 27.6$_{\pm 12.8}$ & 60.9$_{\pm 8.3}$ & 48.4$_{\pm 4.1}$ & 35.1$_{\pm 8.1}$ & 66.4$_{\pm 5.2}$ & 44.8$_{\pm 2.5}$ & 47.2$_{\pm 2.7}$ \\
\dCK & 75.6$_{\pm 2.0}$ & 32.0$_{\pm 16.1}$ & 66.8$_{\pm 8.2}$ & 53.6$_{\pm 4.3}$ & 36.1$_{\pm 10.3}$ & 60.8$_{\pm 3.1}$ & 47.1$_{\pm 4.3}$ & 51.4$_{\pm 6.7}$ \\
\dCL & 77.8$_{\pm 3.0}$ & 34.6$_{\pm 18.9}$ & 72.2$_{\pm 7.1}$ & 61.8$_{\pm 2.9}$ & 45.6$_{\pm 5.2}$ & 59.1$_{\pm 2.4}$ & 48.6$_{\pm 5.8}$ & 51.3$_{\pm 8.0}$ \\
    \midrule
    L50C-PCFG & \textbf{75.2}$_{\pm 2.6}$ & 32.1$_{\pm 17.1}$ & 67.1$_{\pm 8.5}$ & 54.3$_{\pm 3.3}$ & 39.6$_{\pm 8.1}$ & 59.2$_{\pm 2.7}$ & 48.0$_{\pm 4.4}$ & \textbf{51.6}$_{\pm 6.3}$  \\
    L40C-PCFG & 75.0$_{\pm 3.1}$ & 31.9$_{\pm 16.2}$ & 67.2$_{\pm 8.5}$ & 54.6$_{\pm 3.9}$ & 39.7$_{\pm 7.8}$ & 59.4$_{\pm 2.6}$ & 47.8$_{\pm 4.4}$ & 51.3$_{\pm 6.1}$ \\
    L30C-PCFG & 73.0$_{\pm 2.6}$ & 31.6$_{\pm 16.5}$ & 66.4$_{\pm 7.8}$ & 55.2$_{\pm 1.9}$ & 40.2$_{\pm 7.4}$ & 59.7$_{\pm 4.2}$ & 47.1$_{\pm 4.6}$ & 50.5$_{\pm 6.3}$ \\
    L20C-PCFG & 69.2$_{\pm 2.8}$ & 25.9$_{\pm 17.3}$ & 63.8$_{\pm 5.7}$ & 52.7$_{\pm 0.8}$ & 34.0$_{\pm 9.2}$ & 52.2$_{\pm 1.5}$ & 43.7$_{\pm 4.7}$ & 47.1$_{\pm 6.6}$ \\
    L10C-PCFG & 63.3$_{\pm 1.8}$ & 25.5$_{\pm 23.5}$ & 53.7$_{\pm 6.7}$ & 36.2$_{\pm 7.9}$ & 28.2$_{\pm 8.9}$ & 40.2$_{\pm 3.1}$ & 38.3$_{\pm 6.2}$ & 42.8$_{\pm 8.9}$ \\
    % \midrule
    % C-PCFG & 75.0$_{\pm 3.1}$ & 31.9$_{\pm 16.2}$ & 67.2$_{\pm 8.5}$ & 54.6$_{\pm 3.9}$ & 39.7$_{\pm 7.8}$ & 59.4$_{\pm 2.6}$ & 47.8$_{\pm 4.4}$ & 51.3$_{\pm 6.1}$ \\
    % w/ shared \dmonofont{R} & 73.9$_{\pm 1.2}$ & 30.5$_{\pm 11.8}$ & 71.8$_{\pm 1.2}$ & 52.6$_{\pm 2.7}$ & 38.7$_{\pm 4.5}$ & 58.1$_{\pm 1.7}$ & 47.6$_{\pm 2.8}$ & 50.7$_{\pm 4.3}$ \\
    % w/ shared \dmonofont{N} & 74.5$_{\pm 2.0}$ & 38.1$_{\pm 20.4}$ & 70.9$_{\pm 0.8}$ & 56.3$_{\pm 4.7}$ & \textbf{45.4}$_{\pm 9.4}$ & 57.7$_{\pm 2.9}$ & \textbf{49.5}$_{\pm 5.7}$ & 53.1$_{\pm 7.0}$ \\
    % w/ shared \dmonofont{P} & 74.4$_{\pm 0.6}$ & 24.2$_{\pm 4.6}$ & \textbf{74.0}$_{\pm 1.4}$ & 56.8$_{\pm 2.2}$ & 41.3$_{\pm 5.3}$ & \textbf{60.0}$_{\pm 2.5}$ & 46.4$_{\pm 0.5}$ & 48.6$_{\pm 1.4}$ \\
    \bottomrule
\end{tabular}}}}
\caption{\label{tab:test_brown}
Recall on six frequent constituent labels (NP, VP, PP, SBAR, ADJP, ADVP) in the Brown test data, corpus-level F1 (C-F1), and sentence-level F1 (S-F1) results.
The best mean number in each column is in bold.
% $\dagger$ denotes results reported by~\citet{kim-etal-2019-compound}.
L\# indicates that the models are trained on sentences no longer than \# tokens.
}
\end{table*}

\section{Results and discussion}\label{sec:ablation}

% We analyze C-PCFGs on WSJ in Section~\ref{sec:exp_main}-\ref{sec:exp_arch}.
% We perform a quantitative analysis of the correlation between model performance and data distribution in Section~\ref{sec:exp_main}.
% Section~\ref{sec:exp_eff} concerns data efficiency of C-PCFGs and
% presents a length generalization test of C-PCFGs.
% Section~\ref{sec:exp_arch} focuses on the role of sentence embeddings (i.e., the latent vector $\rvz$) in C-PCFGs.
% In Section~\ref{sec:exp_mul} we conduct multilingual evaluation of C-PCFGs.

We primarily perform model and result analysis on English treebanks, including WSJ, Brown, and Enweb (see Table~\ref{tab:test_wsj}-\ref{tab:test_enweb}).

\subsection{Main results}\label{sec:exp_main}

\paragraph{C-PCFGs achieve the highest test performance.}
We compare C-PCFGs against three trivial baselines (left- / right-branching model and random trees)  and a neural PCFG model.
%In the following discussion, We focus on sentence-level F1 for a consistent comparison with~\citet{kim-etal-2019-compound}.
In short, C-PCFGs beats all the baselines in terms of the \emph{maximum} corpus- and sentence-level F1 (see Table~\ref{tab:test_wsj}--\ref{tab:test_enweb}).
Notably, our re-implementation of C-PCFGs reaches the highest sentence-level F1 on WSJ, slightly outperforming the model of~\citet{kim-etal-2019-compound} by 0.5\% S-F1.
While in terms of the mean S-F1, the right-branching baseline outperforms C-PCFGs on Enweb (e.g., +3\% S-F1), considering the high variance (i.e., 5.5\% std), C-PCFGs still have great potential for surpassing the baseline (see Table~\ref{tab:test_enweb}).
The reason that C-PCFGs underperform on Enweb may be because Enweb consists of web text, which tends to be more noisy and informal than WSJ and Brown sentences.

\paragraph{C-PCFGs beat N-PCFGs on all the subdomains of Brown and Enweb.} In Table~\ref{tab:test_brown} and~\ref{tab:test_enweb}, we also present model performance on the subdomains of Brown and Enweb, respectively. Again, C-PCFGs perform best in terms of both corpus- and sentence-level F1 (also see Figure~\ref{fig:ood-subdomain} for an illustration).

\paragraph{Verb phrases are the hardest phrases.}
To give an in-depth analysis of the model gains, we present recall numbers on the top six frequent constituent labels in the test data (NP, VP, PP, SBAR, ADJP, and ADVP).
Unsurprisingly, C-PCFGs achieve the best recall for most labels (i.e., 4 out of 6 constituent labels on all English test sets).
Interestingly, the right-branching baseline always outperforms C-PCFGs on verb phrases (VP) and subordinate clauses (SBAR) (e.g., +30.8\%, +33.1\%, and +41.3\% recall on VPs of WSJ, Brown, and Enweb, respectively), presumably because VPs and SBARs are longer than the rest and involve more complex linguistic structures. On the other hand, this shows that there remains ample room for improvement on VPs and SBARs.

% However, on verb phrases (VPs) they fall far behind the right-branching baseline  (-30.8\% recall),
% presumably because VPs are longer and involve more complex linguistic structures.
\paragraph{C-PCFGs are accurate on short noun phrases.}
We further plot the distributions of the six phrasal labels across constituent lengths (see Figure~\ref{fig:label_dist_per_length}).
We find that VPs are nearly uniformly distributed over different constituent lengths.
In contrast, noun phrases (NPs) concentrate on short constituents.
Take WSJ, NPs account for 61\% of short constituents that have less than 6 tokens and cover 51\% of total constituents (see Figure~\ref{fig:lspan-label-wsj-test}).
We further visualize model performance in terms of recall across constituent lengths (see Figure~\ref{fig:lspan-recall-wsj-test}).
Clearly, short constituents make up a large proportion of total constituents and have a higher chance of being correctly recognized.
Thus, we conclude that C-PCFGs can recognize local and short constituents with high accuracy but struggle with long constituents, such as VPs and SBARs.

\begin{table*}[t]\small
\centering
{\setlength{\tabcolsep}{.65em}
\makebox[\linewidth]{\resizebox{\linewidth}{!}{%
\begin{tabular}{rllllllll}
    \toprule
    \multicolumn{1}{r}{\textbf{Model}} &
    \multicolumn{1}{l}{\textbf{NP}} & 
    \multicolumn{1}{l}{\textbf{VP}} & 
    \multicolumn{1}{l}{\textbf{PP}} & 
    \multicolumn{1}{l}{\textbf{SBAR}} & 
    \multicolumn{1}{l}{\textbf{ADJP}} & 
    \multicolumn{1}{l}{\textbf{ADVP}} & 
    \multicolumn{1}{l}{\textbf{C-F1}} & 
    \multicolumn{1}{l}{\textbf{S-F1}} \\
    \midrule
    Left Branching & \phantom{0}9.9 & \phantom{0}0.9 & \phantom{0}3.4 & 10.1 & \phantom{0}3.9 & 11.2 & \phantom{0}5.8 & 10.9 \\
    Right Branching & 27.1 & \textbf{66.3} & 41.6 & \textbf{59.3} & 30.9 & 29.8 & 38.3 & \textbf{45.9} \\
    Random Trees & 25.1$_{\pm 0.2}$ & 14.7$_{\pm 0.3}$ & 21.6$_{\pm 1.1}$ & 13.0$_{\pm 1.0}$ & 22.1$_{\pm 1.9}$ & 32.9$_{\pm 2.0}$ & 16.8$_{\pm 0.2}$ & 23.1$_{\pm 0.3}$ \\
    \midrule
    N-PCFG & 53.4$_{\pm 4.1}$ & 6.0$_{\pm 3.3}$ & 27.1$_{\pm 22.0}$ & 36.0$_{\pm 10.8}$ & 17.9$_{\pm 8.9}$ & 26.9$_{\pm 8.0}$ & 26.4$_{\pm 5.1}$ & 30.1$_{\pm 3.9}$ \\
    \cdashlinelr{2-9}
    \multicolumn{9}{c}{Per-domain Performance of N-PCFG} \\
    \cdashlinelr{2-9}
\dWEBLOG & 49.5$_{\pm 5.3}$ & 5.5$_{\pm 2.6}$ & 23.4$_{\pm 20.0}$ & 29.5$_{\pm 10.1}$ & 11.9$_{\pm 11.6}$ & 33.9$_{\pm 9.5}$ & 25.2$_{\pm 5.7}$ & 27.2$_{\pm 5.4}$ \\
\dANSWERS & 58.1$_{\pm 5.1}$ & 7.5$_{\pm 4.4}$ & 31.6$_{\pm 24.3}$ & 39.2$_{\pm 10.4}$ & 17.9$_{\pm 11.0}$ & 23.6$_{\pm 11.2}$ & 26.9$_{\pm 5.1}$ & 29.2$_{\pm 4.9}$ \\
\dEMAIL & 50.6$_{\pm 3.6}$ & 5.4$_{\pm 3.2}$ & 25.7$_{\pm 22.2}$ & 33.4$_{\pm 10.8}$ & 9.5$_{\pm 2.9}$ & 23.4$_{\pm 5.4}$ & 24.9$_{\pm 4.8}$ & 30.8$_{\pm 2.8}$ \\
\dNEWSGROUP & 50.1$_{\pm 4.1}$ & 5.5$_{\pm 2.6}$ & 26.8$_{\pm 22.3}$ & 35.6$_{\pm 10.6}$ & 15.7$_{\pm 9.2}$ & 33.3$_{\pm 9.5}$ & 26.8$_{\pm 5.7}$ & 27.8$_{\pm 4.2}$ \\
\dREVIEWS & 59.6$_{\pm 4.0}$ & 5.4$_{\pm 3.3}$ & 28.6$_{\pm 23.1}$ & 40.6$_{\pm 13.1}$ & 24.1$_{\pm 8.5}$ & 28.2$_{\pm 7.2}$ & 28.0$_{\pm 4.8}$ & 32.8$_{\pm 4.2}$ \\
    \cdashlinelr{1-9} 
    C-PCFG & 62.1$_{\pm 3.0}$ & 25.0$_{\pm 10.4}$ & 52.7$_{\pm 14.0}$ & 53.6$_{\pm 2.3}$ & 33.0$_{\pm 7.2}$ & 47.4$_{\pm 10.9}$ & 39.2$_{\pm 4.8}$ & 42.9$_{\pm 5.5}$ \\
    w/ shared \dmonofont{R} & 59.3$_{\pm 6.3}$ & 12.9$_{\pm 6.1}$ & 52.0$_{\pm 12.0}$ & 46.7$_{\pm 3.5}$ & 25.8$_{\pm 4.8}$ & {49.5}$_{\pm 4.0}$ & 34.7$_{\pm 4.8}$ & 38.7$_{\pm 4.8}$ \\
    w/ shared \dmonofont{N} & 62.1$_{\pm 4.0}$ & 23.2$_{\pm 9.0}$ & 59.1$_{\pm 7.0}$ & 46.7$_{\pm 4.6}$ & 30.9$_{\pm 6.0}$ & 51.5$_{\pm 9.0}$ & 39.0$_{\pm 4.1}$ & 42.6$_{\pm 4.0}$ \\
    w/ shared \dmonofont{P} & 58.5$_{\pm 7.3}$ & 11.4$_{\pm 9.0}$ & 49.4$_{\pm 17.0}$ & 46.4$_{\pm 5.2}$ & 24.0$_{\pm 11.0}$ & 48.8$_{\pm 15.4}$ & 33.5$_{\pm 6.8}$ & 36.8$_{\pm 6.2}$ \\
    \cdashlinelr{2-9}
    \multicolumn{9}{c}{Per-domain Performance of C-PCFG} \\
    \cdashlinelr{2-9}
\dWEBLOG & 60.8$_{\pm 3.1}$ & 20.8$_{\pm 8.3}$ & 51.0$_{\pm 9.3}$ & 45.4$_{\pm 4.0}$ & 26.5$_{\pm 9.8}$ & 58.0$_{\pm 9.8}$ & 38.2$_{\pm 3.6}$ & 41.5$_{\pm 5.0}$ \\
\dANSWERS & 67.2$_{\pm 3.7}$ & 29.4$_{\pm 10.6}$ & 59.4$_{\pm 13.7}$ & 54.3$_{\pm 2.4}$ & 36.9$_{\pm 9.8}$ & 48.9$_{\pm 11.6}$ & 41.4$_{\pm 5.0}$ & 45.0$_{\pm 5.4}$ \\
\dEMAIL & 59.5$_{\pm 2.1}$ & 22.4$_{\pm 10.7}$ & 49.6$_{\pm 14.1}$ & 51.8$_{\pm 3.2}$ & 24.6$_{\pm 4.2}$ & 44.5$_{\pm 10.3}$ & 37.3$_{\pm 4.6}$ & 41.5$_{\pm 4.8}$ \\
\dNEWSGROUP & 60.0$_{\pm 2.8}$ & 22.5$_{\pm 8.8}$ & 52.2$_{\pm 10.6}$ & 48.5$_{\pm 4.8}$ & 25.2$_{\pm 5.3}$ & 40.5$_{\pm 8.2}$ & 38.8$_{\pm 4.0}$ & 40.1$_{\pm 5.0}$ \\
\dREVIEWS & 67.9$_{\pm 2.4}$ & 28.9$_{\pm 12.5}$ & 57.1$_{\pm 15.0}$ & 61.1$_{\pm 3.2}$ & 38.0$_{\pm 2.8}$ & 48.4$_{\pm 12.0}$ & 42.4$_{\pm 5.2}$ & 47.0$_{\pm 4.9}$ \\
    \midrule
    L50C-PCFG & \textbf{64.3}$_{\pm 2.7}$ & 25.7$_{\pm 10.9}$ & \textbf{56.2}$_{\pm 11.6}$ & 53.2$_{\pm 1.5}$ & \textbf{35.3}$_{\pm 7.4}$ & \textbf{52.4}$_{\pm 10.1}$ & \textbf{40.6}$_{\pm 4.5}$ & 44.0$_{\pm 5.2}$ \\
    L40C-PCFG & 62.1$_{\pm 3.0}$ & 25.0$_{\pm 10.4}$ & 52.7$_{\pm 14.0}$ & 53.6$_{\pm 2.3}$ & 33.0$_{\pm 7.2}$ & 47.4$_{\pm 10.9}$ & 39.2$_{\pm 4.8}$ & 42.9$_{\pm 5.5}$ \\
    L30C-PCFG & 61.6$_{\pm 2.1}$ & 21.7$_{\pm 10.1}$ & 49.0$_{\pm 15.7}$ & 49.8$_{\pm 2.5}$ & 26.8$_{\pm 10.2}$ & 44.5$_{\pm 16.5}$ & 37.2$_{\pm 4.8}$ & 41.0$_{\pm 5.2}$ \\
    L20C-PCFG & 59.0$_{\pm 4.9}$ & 21.7$_{\pm 8.5}$ & 43.9$_{\pm 11.5}$ & 44.8$_{\pm 3.0}$ & 28.0$_{\pm 4.8}$ & 42.8$_{\pm 13.5}$ & 35.3$_{\pm 3.8}$ & 39.7$_{\pm 4.5}$ \\
    L10C-PCFG & 56.4$_{\pm 2.2}$ & 24.6$_{\pm 9.6}$ & 33.0$_{\pm 4.9}$ & 24.1$_{\pm 4.3}$ & 24.2$_{\pm 2.3}$ & 29.2$_{\pm 2.8}$ & 31.5$_{\pm 3.2}$ & 37.5$_{\pm 2.9}$ \\
    % \midrule
    % C-PCFG & 62.1$_{\pm 3.0}$ & 25.0$_{\pm 10.4}$ & 52.7$_{\pm 14.0}$ & 53.6$_{\pm 2.3}$ & 33.0$_{\pm 7.2}$ & 47.4$_{\pm 10.9}$ & 39.2$_{\pm 4.8}$ & 42.9$_{\pm 5.5}$ \\
    % w/ shared \dmonofont{R} & 59.3$_{\pm 6.3}$ & 12.9$_{\pm 6.1}$ & 52.0$_{\pm 12.0}$ & 46.7$_{\pm 3.5}$ & 25.8$_{\pm 4.8}$ & {49.5}$_{\pm 4.0}$ & 34.7$_{\pm 4.8}$ & 38.7$_{\pm 4.8}$ \\
    % w/ shared \dmonofont{N} & 62.1$_{\pm 4.0}$ & 23.2$_{\pm 9.0}$ & 59.1$_{\pm 7.0}$ & 46.7$_{\pm 4.6}$ & 30.9$_{\pm 6.0}$ & 51.5$_{\pm 9.0}$ & 39.0$_{\pm 4.1}$ & 42.6$_{\pm 4.0}$ \\
    % w/ shared \dmonofont{P} & 58.5$_{\pm 7.3}$ & 11.4$_{\pm 9.0}$ & 49.4$_{\pm 17.0}$ & 46.4$_{\pm 5.2}$ & 24.0$_{\pm 11.0}$ & 48.8$_{\pm 15.4}$ & 33.5$_{\pm 6.8}$ & 36.8$_{\pm 6.2}$ \\
    \bottomrule
\end{tabular}}}}
\caption{\label{tab:test_enweb}
Recall on six frequent constituent labels (NP, VP, PP, SBAR, ADJP, ADVP) in the Enweb test data, corpus-level F1 (C-F1), and sentence-level F1 (S-F1) results.
The best mean number in each column is in bold.
% $\dagger$ denotes results reported by~\citet{kim-etal-2019-compound}.
L\# indicates that the models are trained on sentences no longer than \# tokens.
}
\end{table*}
\begin{figure*}[t!]
\centering
\begin{subfigure}{.99\linewidth}
    \includegraphics[width=1.\linewidth]{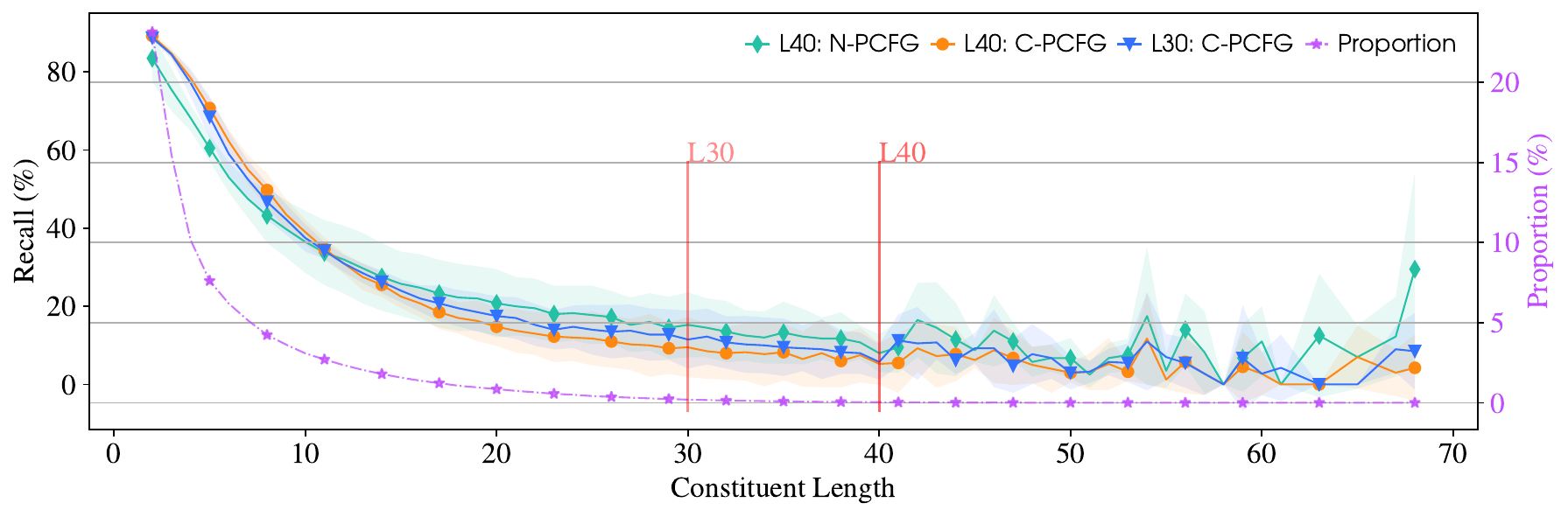}
    \caption{WSJ Training Set.}
    \label{fig:recall-len-wsj-train}
\end{subfigure}
\begin{subfigure}{.99\linewidth}
    \includegraphics[width=1.\linewidth]{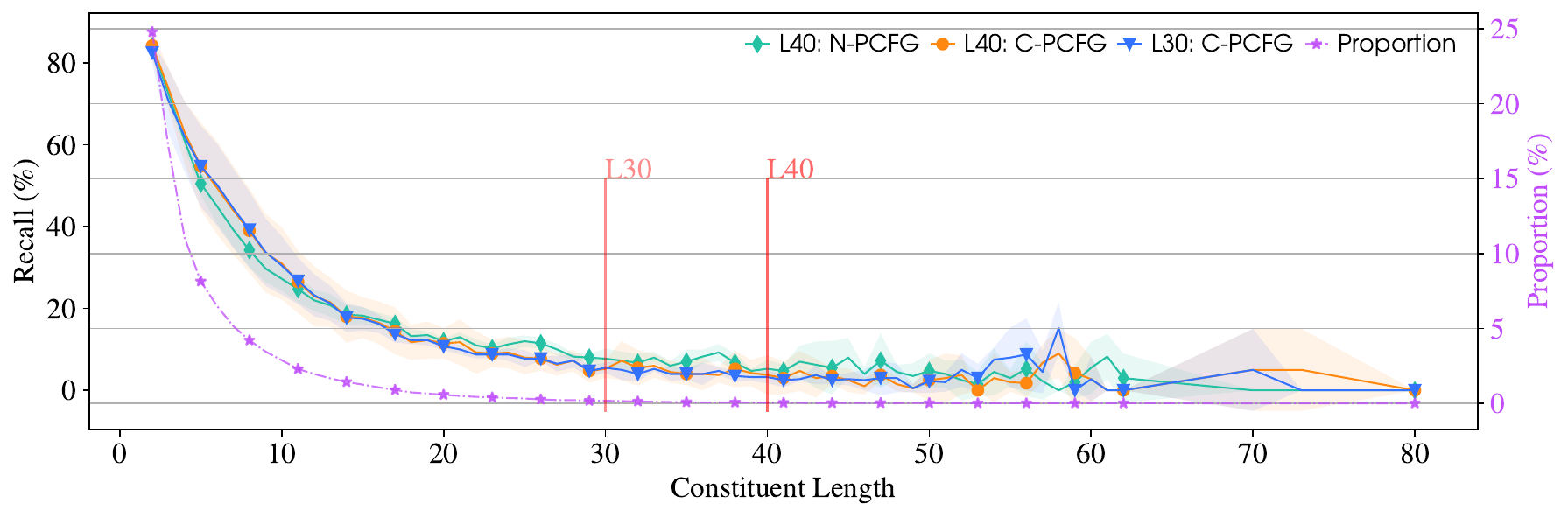}
    \caption{Brown Training Set.}
    \label{fig:recall-len-brown-train}
\end{subfigure}
\begin{subfigure}{.99\linewidth}
    \includegraphics[width=1.\linewidth]{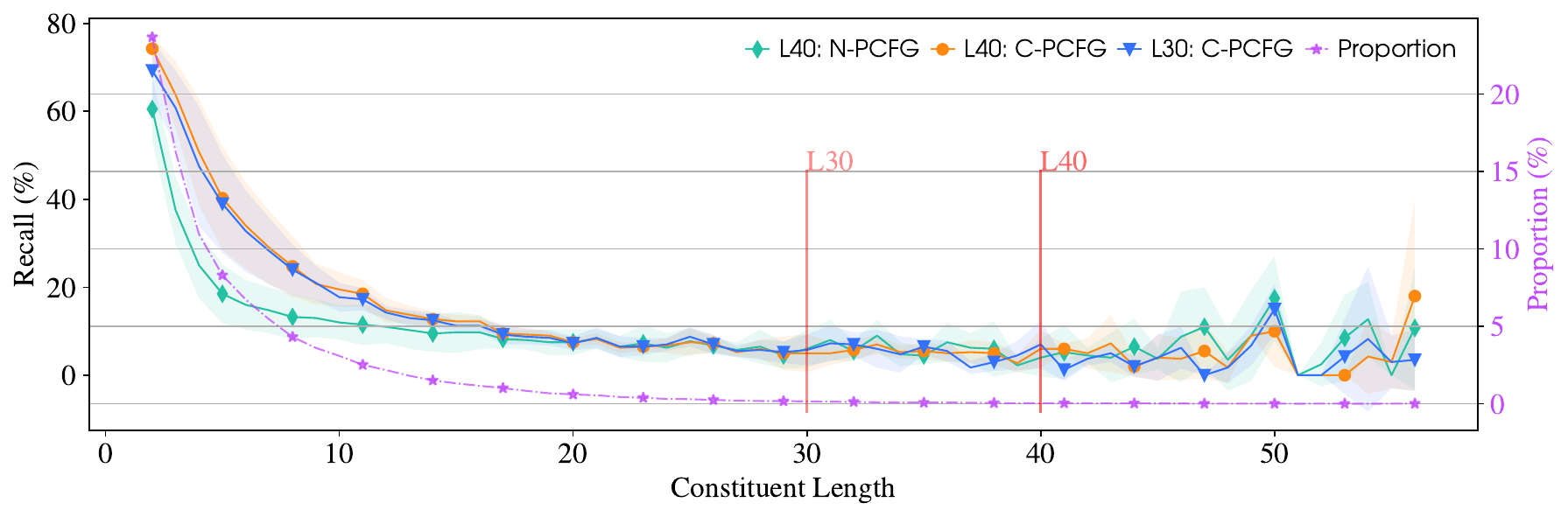}
    \caption{Enweb Training Set.}
    \label{fig:recall-len-enweb-train}
\end{subfigure}
\caption{\label{fig:recall_per_length_train}
F1 numbers broken down by constituent lengths on the WSJ training data.
During training, constituents (sentences) longer than 30 tokens (L30) are unseen to L30C-PCFG and are unseen to L40C-PCFG and L40N-PCFG when longer than 40 tokens (L40).
}
\end{figure*}

\subsection{Data efficiency and length generalization}\label{sec:exp_eff}

A crucial aspect of human languages is their compositionality. %that they exhibit compositional structures.
Humans can derive grammar rules from a few sentences and combine the rules to generate new sentences compositionally.
As C-PCFGs are backed by context-free grammar, % (CFGs),
we hypothesize that C-PCFGs are data-efficient and have good generalizability to unseen sentence/constituent lengths.
%At test time, we evaluate C-PCFGs either on the test data of WSJ (to check data efficiency) or on the training data of WSJ (to perform length generalization test). 
To verify our hypothesis, we design a length-generalization test.
We train C-PCFGs using training sentences of length equal to or below a chosen sentence length.
In our experiments, we choose five sentence lengths (i.e., 10, 20, 30, 40, and 50) and indicate the resulting models by prefixes L10, L20, L30, L40, and L50, respectively (see the last rows of Table~\ref{tab:test_wsj}-\ref{tab:test_enweb}).

\paragraph{Training on more and longer sentences is always helpful.}
We test all the learned models on the corresponding full test sets. In Figure~\ref{fig:efficiency-final}, we illustrate their sentence-level F1 numbers.
Overall, training C-PCFGs on more and longer sentences results in higher F1 numbers.
But using training sentences longer than 40 tokens only results in very small improvements (i.e., +0.2\%, +0.3\%, and +1.1\% F1 on WSJ, Brown, and Enweb, respectively).
%hurt the performance (-0.1\% F1).
Given that more than 96\% of all the test sentences are shorter than 41 tokens,
we conjecture that training sentences shorter than 41 tokens have adequately covered the syntactic phenomena in the test data.
On the other hand, longer sentences have a larger tree space and probably make it harder for C-PCFGs to disambiguate parse trees.

\paragraph{C-PCFGs are data-efficient.}
On all three English treebanks, we find that discarding training sentences longer than 30 tokens only decreases the model performance by less than 2\% S-F1, which suggests that C-PCFGs are data-efficient.
Surprisingly, training C-PCFGs only on sentences shorter than 11 tokens already gives rise to good parsing performance, e.g., L10C-PCFGs achieve 48.2\% S-F1 on WSJ and 42.8\% S-F1 on Brown.
%showing that C-PCFGs are robust to sentence length and can generalize to longer sentences.

\begin{figure*}[t!]
\centering
\begin{subfigure}{.33\linewidth}
    \includegraphics[width=1.\linewidth]{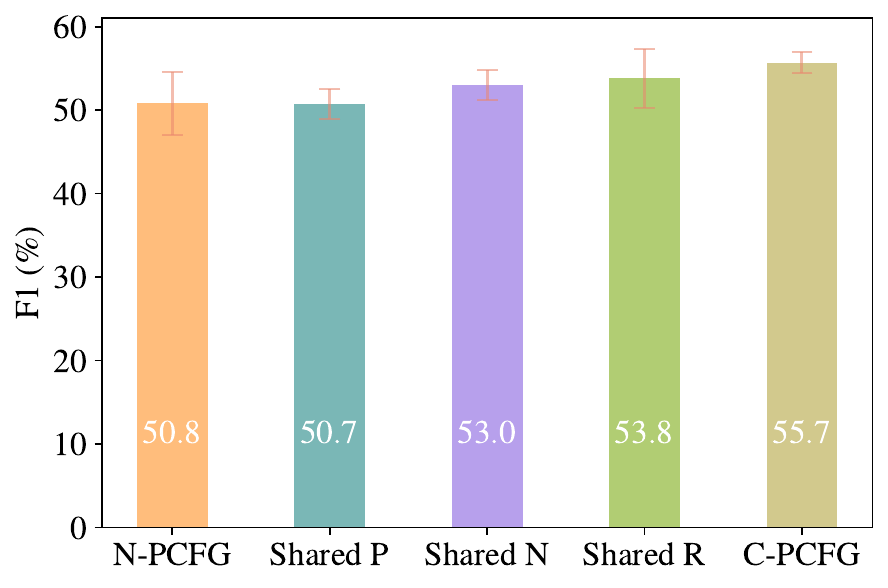}
    \caption{WSJ Test Set.}
    \label{fig:ablation-wsj-test}
\end{subfigure}
\hfill
\begin{subfigure}{.33\linewidth}
    \includegraphics[width=1.\linewidth]{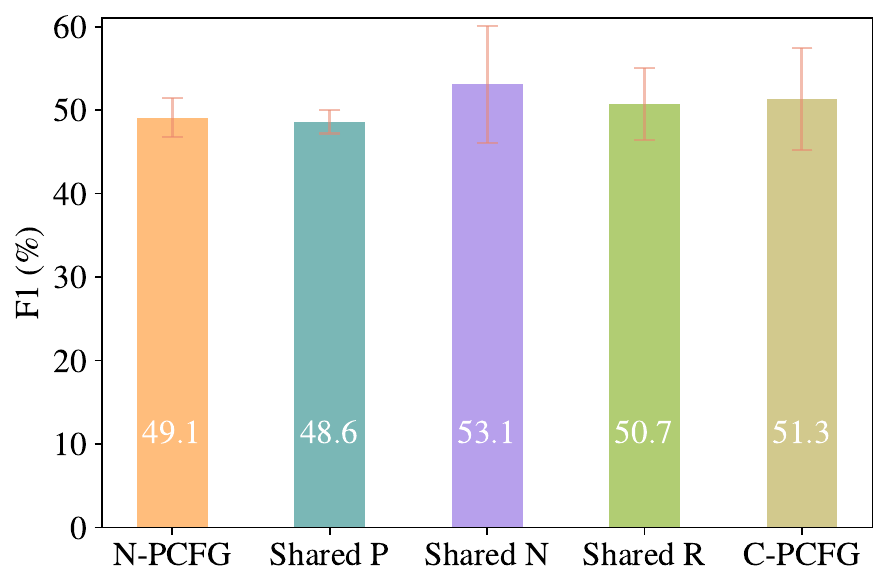}
    \caption{Brown Test Set.}
    \label{fig:ablation-brown-test}
\end{subfigure}%
\hfill
\begin{subfigure}{.33\linewidth}
    \includegraphics[width=1.\linewidth]{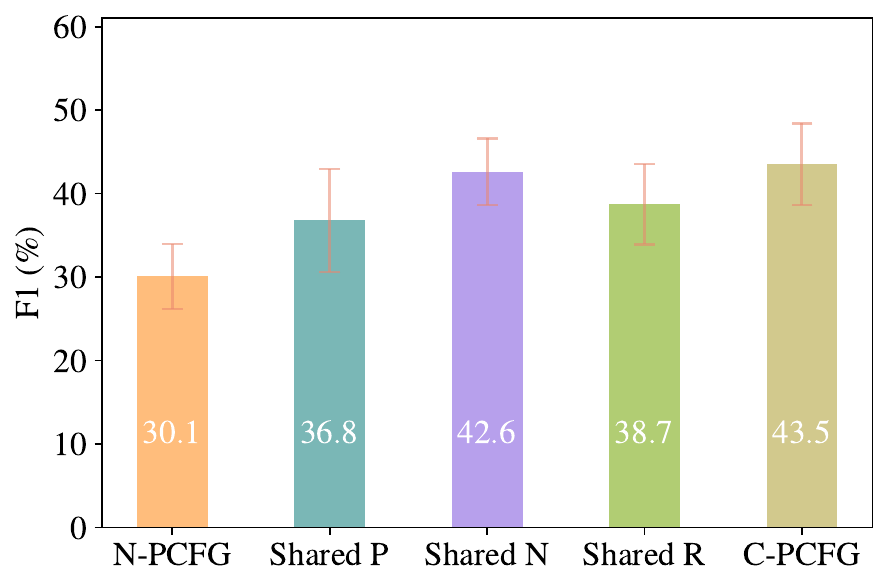}
    \caption{Enweb Test Set.}
    \label{fig:ablation-enweb-test}
\end{subfigure}%
\caption{\label{fig:ablation}
F1 numbers on the three test sets.
Shared P / N / R indicates C-PCFGs that use corpus-level parameters for preterminal / nonterminal / start rules (see Section~\ref{sec:exp_arch}).
}
\end{figure*}

\paragraph{C-PCFGs can generalize to unseen constituent lengths.}
To study the generalizability of C-PCFGs to unseen longer constituents, we conduct a length generalization test.
%(see Appendix~\ref{sec:exp_constituent_len} for details).
%We perform a constituent-length generalization test.
Since the test sets do not have sufficient constituent statistics across constituent lengths, to make our analysis reliable, we instead test C-PCFGs on training sentences and report model performance in terms of recall (see Figure~\ref{fig:recall_per_length_train}).
% across \textit{constituent lengths}
%\footnote{We do not report F1 numbers across sentence lenghts because test sentences may contain constituents which are present in training.} 
In general, F1 numbers become lower as constituent length increases.
This is reasonable because large constituents are merged from small constituents;
errors from small constituents accumulate when composing larger constituents.

To investigate the influence of training data on length generalization, we use sentence lengths 30 and 40 for an illustration and analyze the results from WSJ (see Figure~\ref{fig:recall-len-wsj-train}).
Surprisingly, when tested on constituents longer than 40 tokens, L30C-PCFG shows slightly better performance than L40C-PCFG.
Notably, it consistently outperforms L40C-PCFG on sentences of length between 30 and 40, though L40C-PCFG has been trained on additional sentences of length ranging from 30 to 40.
This finding confirms that C-PCFGs are able to generalize to unseen constituent/sentence lengths, but training on additional longer sentences may hurt generalizability. 

In Figure~\ref{fig:recall_per_length_train}, we also plot the proportion that each constituent length makes up of total constituents.
We find that there are about 6,400 constituents of length between 30 and 40, which account for about 1.1\% of total constituents and give rise to relatively sufficient statistics, thus making our conclusion about the generalizability of L30C-PCFGs reliable.

We further compare C-PCFGs with L40N-PCFGs on WSJ (see Figure~\ref{fig:recall-len-wsj-train}). Surprisingly, L40N-PCFGs demonstrate the best generalizability on long constituents.
A natural question that follows is: \textit{where does the F1 improvement of C-PCFGs over N-PCFGs come from?}
Look at the recall on shorter constituents, clearly, C-PCFGs perform better on constituents that are shorter than 11 tokens,
while L40N-PCFGs consistently outperform C-PCFGs on constituents of length between 11 and 40, and L30C-PCFGs fall in between C-PCFGs and L40N-PCFGs.
This, once again, demonstrates that training C-PCFGs on short sentences may be adequate for good parsing performance and reasonable generalizability, presumably because of the high data efficiency of C-PCFGs.

We have so far focused on result analysis on WSJ. On Brown and Enweb, though we do not observe clear differences among C-PCFGs, L30C-PCFGs, and L40N-PCFGs on long constituents, we do find that, on short constituents, C-PCFGs outperform N-PCFGs and the improvement is especially large on Enweb, e.g., +37.5\% recall for the constituent length of 3.

% not only gives rise to good parsing performance but also brings about decent generalizability.

\begin{table*}[t!]\small
\centering
\vskip -.0in
{\setlength{\tabcolsep}{.5em}
\makebox[\linewidth]{\resizebox{\linewidth}{!}{%
\begin{tabular}{rllllllllll}
    \toprule
    \textbf{Model} &
    \textbf{Chinese}  &
    \textbf{Basque} &
    \textbf{German} &
    \textbf{French} &
    \textbf{Hebrew} &
    \textbf{Hungarian} &
    \textbf{Korean} &
    \textbf{Polish} &
    \textbf{Swedish} &
    \textbf{Mean} 
    \\ 
    \midrule
    Left Branching & \phantom{0}7.2 & 17.9 & 10.0 & \phantom{0}5.7 & \phantom{0}8.5 & 13.3 & 18.5 & 10.9 & \phantom{0}8.4 &  11.2 \\
    Right Branching & 25.5 & 15.4 & 14.7 & 26.4 & 30.0 & 12.7 & 19.2 & \textbf{34.2} & \textbf{30.4} & 23.2 \\
    Random Trees & 15.2 & 19.5 & 13.9 & 16.2 & 19.7 & 14.1 & 22.2 & 21.4 & 16.4 & 17.6 \\
    N-PCFG & 30.1$_{\pm 4.6}$ & \textbf{30.2}$_{\pm 0.9}$ & \textbf{37.8}$_{\pm 1.7}$ & \textbf{42.2}$_{\pm 1.4}$ & \textbf{41.0}$_{\pm 0.6}$ & 37.9$_{\pm 0.8}$ & 25.7$_{\pm 2.8}$ & 31.7$_{\pm 1.8}$ & 14.5$_{\pm 12.7}$ & 32.3 \\
    C-PCFG & \textbf{35.1}$_{\pm 6.1}$ & 27.9$_{\pm 2.0}$ & 37.3$_{\pm 1.8}$ & 40.5$_{\pm 0.8}$ & 39.2$_{\pm 1.2}$ & \textbf{38.3}$_{\pm 0.7}$ & \textbf{27.7}$_{\pm 2.8}$ & 32.4$_{\pm 1.1}$ & 23.7$_{\pm 14.3}$ & \textbf{33.6} \\
    \bottomrule
\end{tabular}}}}
\caption{\label{tab:miltilingual} Sentence-level F1 numbers on multilingual treebanks.
Similarly to \citet{kim-etal-2019-compound}, we observe that C-PCFGs suffer a huge variance, e.g., on the Chinese and Swedish treebanks.
}
\vskip -.0in
\end{table*}

\subsection{Model ablation}\label{sec:exp_arch}

Our experimental results have so far shown that C-PCFGs improve over N-PCFGs. 
From a modeling perspective, the major difference between C-PCFGs and N-PCFGs is that C-PCFGs use an additional sentence embedding, i.e., the latent variable $\rvz$ (see Section~\ref{sec:background}), to construct a sentence-specific parameterization of PCFGs. Concretely, the sentence embedding is used to compute the probabilities of three types of rules: (1) preterminal rules (P), (2) nonterminal rules (N), and (3) start rules (R).
We would like to know: \textit{out of the three types of rules, which type makes the best use of the sentence embedding?}
To answer this question,
we ablate C-PCFGs by letting C-PCFGs use corpus-level parameters for each of the three rule types, individually. In other words, the parameters of an ablated rule type will be shared across sentences.

Interestingly, on both WSJ and Brown, C-PCFGs degenerate into N-PCFGs when using corpus-level parameters for preterminal rules (see Figure~\ref{fig:ablation-wsj-test} and~\ref{fig:ablation-brown-test}). Similarly, on Enweb, sharing preterminal rule probabilities results in decreased performance (-5.4\% F1). On the whole, sharing the probabilities of preterminal rules leads to the largest reduction in performance. 
This implies that the sentence embedding is most crucial for the parameterization of preterminal rule probabilities,
presumably because the sentence embedding is conducive to deriving the knowledge of part-of-speech tags for preterminal rules, which is further beneficial for inducing larger phrase structures.

\subsection{Multilingual evaluation}\label{sec:exp_mul}

Despite the impressive performance of C-PCFGs on English,
it is still unclear whether C-PCFGs can generalize to other languages.
Thus, we further conduct a multilingual evaluation of C-PCFGs on nine additional languages. In training, we use the hyperparameters of the best-performing C-PCFG on English, without performing additional hyperparameter search on the nine languages.
% i.e., we tune C-PCFGs only on WSJ and use the best configurations on the other treebanks.

In general, C-PCFGs achieve the highest overall mean F1, which is averaged over all nine treebanks,
though they have two fewer winning treebanks than N-PCFGs (see Tablel~\ref{tab:miltilingual}).
Notably, both C-PCFGs and N-PCFGs outperform the trivial baselines by a large margin,
indicating that they are able to generalize to languages beyond English.
On individual treebanks, however, C-PCFGs sometimes underperform the right-branching baseline, e.g., on the Polish and Swedish treebanks, presumably because these languages have rich morphologies.
Thus, we anticipate that encoding the knowledge of morphologies into the sentence embedding will improve C-PCFGs further.
% an improvement from encoding the knowledge of morphologies into the sentence embedding.

\section{Conclusion}

We have conducted an analysis of C-PCFGs from a quantitative perspective.
Our analysis focuses on four aspects of C-PCFGs:
data efficiency, length generalization, the role of the sentence embedding, and multilingual generalization.
We empirically find that C-PCFGs are data-efficient and are able to generalize to unseen constituent/sentence lengths. 
% can learn well only from short sentences and maintain good performance at test time.
Our ablation study suggests that the sentence embedding in C-PCFGs is most crucial for preterminal rules. 
Despite being performant in general, we find that the configurations of the best-performing C-PCFGs on English do not always generalize to morphology-rich languages.

\section*{Acknowledgments}
We would like to thank anonymous reviewers for their 
suggestions and comments. The project was supported by the
European Research Council (ERC Starting Grant BroadSem
678254) and the Dutch National Science Foundation
(NWO VIDI 639.022.518).

\bibliography{anthology,eacl2021}
\bibliographystyle{acl_natbib}

%\appendix

%\section{Constituent length generalization}\label{sec:exp_constituent_len}

\end{document}